%% file: acl.tex
\pdfoutput=1
\documentclass[11pt]{article}

\usepackage{acl}

\usepackage{times}
\usepackage{latexsym}

\usepackage[T1]{fontenc}
\usepackage[utf8]{inputenc}

\usepackage{microtype}

\usepackage{amsmath}
\usepackage{amssymb}
\usepackage{xcolor}
\usepackage{arydshln}
\usepackage{tikz}
\usepackage{pgfplots}
\pgfplotsset{compat=1.17}
\usepackage{subcaption}
\usepackage{mathtools}
\usepackage{multirow}
\usepackage{booktabs}
\usepackage{graphicx}

\title{Attention Mechanism with Energy-Friendly Operations}

\author{Yu Wan$^{a,b}$\thanks{~~Equal contribution.}~\thanks{~Work was done when interning at DAMO Academy, Alibaba Group.}~~~Baosong Yang$^{b*}$~~~Dayiheng Liu$^b$~~~Rong Xiao$^{b \dagger}$\thanks{~~Corresponding author.}~~~Derek F. Wong$^{a}$\\\textbf{Haibo Zhang}$^b$~~~\textbf{Boxing Chen}$^b$~~~\textbf{Lidia S. Chao}$^a$ \\
  $^a$NLP$^2$CT Lab,
  University of Macau\\
  {\tt nlp2ct.ywan@gmail.com, \{derekfw,lidiasc\}@umac.mo} \\
  $^b$Alibaba Group\\
  {\tt \{yangbaosong.ybs,liudayiheng.ldyh,birong.xr,zhanhui.zhb,}\\
  {\tt boxing.cbx\}@alibaba-inc.com}}

\begin{document}
\maketitle
\begin{abstract}

Attention mechanism has become the dominant module in natural language processing models. 
It is computationally intensive and depends on massive power-hungry multiplications.
In this paper, we rethink variants of attention mechanism from the energy consumption aspects. After reaching the conclusion that the energy costs of several energy-friendly operations are far less than their multiplication counterparts, we build a novel attention model by replacing multiplications with either selective operations or additions.
Empirical results on three machine translation tasks demonstrate that the proposed model, against the vanilla one, achieves competitable accuracy while saving 99\% and 66\% energy during alignment calculation and the whole attention procedure.
Code is available at: \href{https://github.com/NLP2CT/E-Att}{https://github.com/NLP2CT/E-Att}.

\end{abstract}

\section{Introduction}

Attention mechanism \citep[ATT,][]{bahdanao2014neural,vaswani2017attention,yang2018modeling} has demonstrated huge success in a variety of natural language processing tasks~\cite{su2018hierarchy,kitaev2018constituency,tan2018deep,yang2019context,devlin2019bert,zhang2020neural}. 
The module learns hidden representations of a sequence by serving each word as a query to attend to all keys in the target sentence, then softly assembling their values. 
It is a de-facto standard to achieve this via performing linear projections and dot products on representations of queries and keys~\cite{vaswani2017attention}, resulting in large amount of multiplications.
In spite of its promising quality, such kind of paradigm may be not the preferred solution from the energy consumption aspect~\cite{horowitz2014computing,raffel2020exploring}.  
How to build a high energy-efficient ATT still remains a great challenge.

\input{table_cmos_energy}
%\textcolor{red}{
Our work starts from in-depth investigations on approaches in ATT context with respect to model compression~\cite{hinton2015distilling,jiao2020tinybert} and complexity optimization~\cite{yang2019convolutional,raganato2020fixed,beltagy2020longformer,tay2021synthesizer}. 
These approaches can potentially alleviate the problem of high energy consumption in ATT. % by reducing demanded computations.
Nevertheless, intentions of all these methods are not exactly from the energy-friendly perspective, thus overlooking the origin of energy consumed, i.e., basic arithmetic operations in electric equipments. 
%\textcolor{red}{Although those approaches can potentially alleviate the problem of high energy consumption in ATT, their intentions do not strictly lie in the energy-friendly perspective.}
Massive multiplications still remain, consuming far more energy than its additive counterpart on modern devices~\citep[Table~\ref{tab:cmos},][]{li2020survey}.

To this end, we propose to approach this problem from a new direction -- replacing massive multiplications in ATT with cheaper operations. 
%\textcolor{red}{Driven by this finding, we attempt to achieve the aim of energy reduction of ATT via a new direction -- replacing massive multiplications in ATT with cheaper operations.}
Concretely, we propose a novel energy-efficient attention mechanism (E-ATT). 
It equips binarized selective operations instead of linear projections over input hidden states, and measures attentive scores using $L_1$ distance rather than dot-product.
Consequently, E-ATT abandons most of multiplications to reach the goal of energy cost reduction.

We examine our method with \textsc{Transformer} model~\cite{vaswani2017attention}, and conduct experiments on three machine translation tasks.
Compared with conventional ATT, our E-ATT can save more than 99\% energy of the vanilla alignment calculation and around 66\% energy of the whole attention model. In the meanwhile, our models yield acceptable translation qualities across language pairs.
Extensive analyses also demonstrate that E-ATT can functionally model semantic alignments without using multiplications.

\section{Preliminary}
\paragraph{Conventional Attention Mechanism}
Given input representations $\mathbf{X} \in \mathbb{R}^{l_1 \times d}$ and $\mathbf{Y} \in \mathbb{R}^{l_2 \times d}$ with $l_1$, $l_2$ being sequence length, and $d$ is the input dimensionality.
Note $l_1$ and $l_2$ may be equal for self-attention pattern, and represent lengths of target and source sequence in cross-attention. 
ATT first projects the inputs into three representations\footnote{For simplicity, we omit the bias in linear projections, as well as splits and concatenations in multi-head mechanism.}:
\begin{align}
    \mathbf{Q} & = \mathbf{X} \mathbf{W}_Q , ~~~ [\mathbf{K}; \mathbf{V}] = \mathbf{Y} [\mathbf{W}_K; \mathbf{W}_V], \label{equ.QKVTransition}
    % Remove equations
    % \hat{\mathbf{Q}} & = \textit{SplitHead} (\mathbf{Q}) \in \mathbb{R}^{h \times l_1 \times k_h}, \label{equ.Qsplithead} \\
    % \hat{\mathbf{K}}; \hat{\mathbf{V}} & = \textit{SplitHead} ([\mathbf{K}, \mathbf{V}]) \in \mathbb{R}^{h \times l_2 \times k_h}, \label{equ.KVsplithead}
\end{align}
% where $\mathbf{W}_Q, \mathbf{W}_K \in \textcolor{red}{\mathbb{R}^{d \times d}}, \mathbf{W}_W \in \mathbb{R}^{d \times d}$ are parameters, $k$ and $k_h$ are the dimensionalities for linear transition and each head, $\mathbf{Q}$, $\mathbf{K}$ and $\mathbf{V}$ are query, key and value representations, respectively.
where $\{\mathbf{W}_Q, \mathbf{W}_K, \mathbf{W}_V\} \in \mathbb{R}^{d \times d}$ are trainable parameters. $\mathbf{Q} \in \mathbb{R}^{l_1 \times d}$, $\{\mathbf{K}, \mathbf{V}\} \in \mathbb{R}^{l_2 \times d}$ are query, key and value representations, respectively.
% For the $m$-th head, the attention alignment is calculated by obtaining the logits with dot-product multiplication, following by softmax activation:
The attention alignment is calculated with dot-product multiplication and softmax activation:
\begin{align}
    % \mathbf{A}_{ij}^m \propto \exp(\frac{\hat{\mathbf{Q}}_{i}^m \hat{\mathbf{K}}_{j}^{m\top} }{\sqrt{k_h}}) \in \mathbb{R}^{l_1 \times l_2}.
    % \label{equ.DotProductAttnScore}
    \mathbf{M}_{ij} \propto \exp(\frac{\mathbf{Q}_{i} {\mathbf{K}}_{j}^{\top} }{\sqrt{d}}) \in \mathbb{R}^{l_1 \times l_2}.
    \label{equ.DotProductAttnScore}
\end{align}

% Then, the output is derived by multiplying attention weights with value representation $\hat{\mathbf{V}}$, concatenating heads and additional linear projection:
Then, the output is derived by multiplying attention weights with value representation $\hat{\mathbf{V}}$:
\begin{align}
    \mathbf{O} & = \mathbf{M} {\mathbf{V}} \in \mathbb{R}^{l_1 \times d}. \label{equ.VMatrixMul}
    % \hat{\mathbf{O}}^m & = \mathbf{A}^m \hat{\mathbf{V}}^m \in \mathbb{R}^{l_1 \times h_k}, \label{equ.VMatrixMul} \\
    % \mathbf{O} & = \textit{ConcatHead} (\hat{\mathbf{O}}) \in \mathbb{R}^{l_1 \times d}, \label{equ.ConcatHead}
\end{align}
As seen, matrix multiplications are massively exploited into conventional ATT.

\paragraph{Related Work}
Several related approaches potentially alleviate the power-hungry drawback of ATT.
One direction relies on model compression by pruning redundant parameters~\cite{denton2014exploiting,wang2016cnnpack,zhuang2018discrimination} or distilling the learned knowledge from a large model to a smaller one~\cite{hinton2015distilling,yim2017gift}, which still maintains multiplicative operations.
Another direction aims at reducing the computational complexity of attention module, e.g. linearly projecting input~\citep[Dense,][]{tay2021synthesizer}, or randomly initializing and training attention weights~\citep[RandInit,][]{tay2021synthesizer}. 
To give a full comparison of energy consumption of these approaches, we conduct the number of multiplicative operations and energy costs across modules in Table~\ref{tab:stats_model_operation}.
As seen, vanilla ATT~\cite{vaswani2017attention} involves the most multiplicative operations, and requires the most energy than other methods.
Although Dense and RandInit significantly reduce the energy consumption,  \newcite{tay2021synthesizer} point out that these approaches fail to be employed into cross-attention networks, since neither linear transition nor randomly initialized matrix is able to exactly model alignment information across languages. 

\input{table_calls_of_operators}
%We argue that the main reason stems from the mechanism of these modules.  Since attention weights are obtained from the linear transition of  representations, or randomly initialized parametric matrix, these modules can not strongly contribute aligned information across languages. 

\section{Energy-Efficient Attention Mechanism}
% In this section, we describe our E-ATT including the aspects of 1) discrete selective operations and 2) $L_1$ distance measurements to reduce the multiplicative operations.

In this section, we describe E-ATT by pertinently reducing the multiplicative operations of ATT, including selective operation and $L_1$ distance.

\subsection{Feature Selection with Discreteness}
Since the linear transitions of queries and keys (Equation~\ref{equ.QKVTransition}) involve massive multiplications within conventional ATT, we propose to modify them with binarized quantization~\cite{liu2018binarized,qin2020binary}.
Concretely, the inputs $\mathbf{X}$ and $\mathbf{Y}$ are turned into discrete value with a threshold function $f(\cdot)$:
\begin{align}
    f(x) = 
        \begin{cases}
            1 & x > \tau, \\
            0 & \text{otherwise,}
        \end{cases}
    \label{Equ.binary_function}
\end{align}
where $\tau$ and $d$ are threshold and hidden size, respectively.
The derived representations $\tilde{\mathbf{X}} = f(\mathbf{X})$ and $\tilde{\mathbf{Y}} = f(\mathbf{Y})$  contain discrete features composing of zeros and ones.
Since this procedure is undifferentiable, we need to predefine a pattern of gradient calculation for $\mathbf{X}$ when receiving back-propagated gradient $\mathbf{Z}$. 
\newcite{wu2018spatio} pointed out that, when simulating the back-propagated progress across discrete activations, those patterns which peak at the medium of domain reveal better training stabilization and model performance. We thus use a modified Gaussian function during back-propagation following  \newcite{wu2018spatio}:
% Inspired by recent work~\cite{wu2018spatio}, here we use a modified Gaussian function during back-propagation:
\begin{align}
    \nabla \mathbf{X} = \sqrt{\frac{2}{\pi}} e ^ {-2 (\mathbf{Z} - \tau)^2},
\end{align}
and the same procedure is applied for $\mathbf{Y}$.
Then given parameters $\mathbf{W}_Q, \mathbf{W}_K \in \mathbb{R}^{d \times d}$, we derive query and key representations $\mathbf{Q}, \mathbf{K}$ by applying masked selection function:
\begin{align}
    \tilde{\mathbf{Q}} & = g (\tilde{\mathbf{X}}, \tilde{\mathbf{W}_Q)} \in \mathbb{R}^{l_1 \times d \times d}, \\
    \tilde{\mathbf{K}} & = g (\tilde{\mathbf{Y}}, \tilde{\mathbf{W}_K)} \in \mathbb{R}^{l_2 \times d \times d}, \\
    \mathbf{Q} & = \sum_{i=1}^{d} \tilde{\mathbf{Q}}_{\cdot, i, \cdot};~~~\mathbf{K} = \sum_{i=1}^{d} \tilde{\mathbf{K}}_{\cdot, i, \cdot}, 
\end{align}
where $\tilde{\mathbf{W}_Q} \in \mathbb{R}^{l_1 \times d \times d}$ and $\tilde{\mathbf{W}_K} \in \mathbb{R}^{l_2 \times d \times d}$ are derived by tiling $\mathbf{W}_Q, \mathbf{W}_K$ with $l_1$ and $l_2$ times, respectively.  $g(\cdot, \cdot)$ represents indexed feature selection defined as follows:
\begin{align}
    g(\mathbf{U}, \mathbf{P}) = 
        \begin{cases}
            \mathbf{U}_{i,j,\cdot} & \mathbf{P}_{i,j} = 1, \\
            \mathbf{0} & \text{otherwise.}
        \end{cases}
\end{align}

\subsection{Pairwise $L_1$ Distance}
As the dot-product multiplication can be viewed as similarity calculation between $\mathbf{Q}$ and $\mathbf{K}$, we argue that other similarity estimation methods can play this role as well. 
Accordingly, we further propose to use pairwise $L_1$ distance instead, which does not require any multiplication.
Attention score calculation in Equation~\ref{equ.DotProductAttnScore} is then modified as:
\begin{align}
    % \mathbf{A}_{ij}^m \propto \exp(- \frac{|| \hat{\mathbf{Q}}_{i}^m - \hat{\mathbf{K}}_{j}^m ||_1}{\sqrt{k_h}}),
    \mathbf{M}_{ij} \propto \exp(- \frac{|| {\mathbf{Q}}_{i} - {\mathbf{K}}_{j} ||_1}{\sqrt{d}}),
\end{align}
where $||\cdot||_1$ denotes the $L_1$ norm of inputted vector.
Here we use negative $L_1$ value to ensure that larger distance contributes lower attention score.
% The following computational process is identical to conventional attention networks (Equation~\ref{equ.VMatrixMul}$\sim$\ref{equ.ConcatHead}).
%As E-ATT does not require any multiplication (Table~\ref{tab:stats_model_operation}), our model can significantly reduce the energy consumption to 0.44\% of baseline.

\section{Experiments}
% \subsection{Dataset and Model Setting}

% We choose three machine translation tasks, i.e. IWSLT'15 English-Vietnamese (En-Vi), WMT'14 English-German (En-De) and WMT'17 Chinese-English (Zh-En), to evaluate the effectiveness of our approach.
% We follow the setting of \textsc{Transformer}-\textit{Base}~\cite{vaswani2017attention} for all involved tasks, with model hidden size $d$ as 512, the number of layers in encoder and decoder as 6, and the number of attention head as 8. 

\subsection{Dataset Preprocessing}
In this paper we evaluate our approach with three widely used machine translation datasets: IWSLT'15 English -Vietnamese (En-Vi), WMT'14 English - German (En-De) and WMT'17 Chinese - English (Zh-En).
All datasets are segmented into subwords by byte-pair encoding ~\citep[BPE,][]{sennrich2016neural} with 32k merge operations.
Specially, for the former two tasks, we apply joint BPE for both source and target languages.
% All datasets are modified into truecase format with \texttt{mosesdecoder}\footnote{\href{https://github.com/mosesdecoder/}{https://github.com/mosesdecoder/}} by training truecase models upon train set.
% All datasets are modified into truecase format with \texttt{mosesdecoder}\footnote{\href{https://github.com/mosesdecoder/}{https://github.com/mosesdecoder/}}.

\begin{table}[t]
    \centering
    \begin{tabular}{cccc}
        \toprule
        \textbf{Dataset} & \textbf{Train} & \textbf{Dev} & \textbf{Test} \\
        \midrule
        En-Vi & 13.3K & 1,553 & 1,268 \\
        En-De & 4.50M & 3,000 & 3,003 \\
        Zh-En & 20.6M & 2,002 & 2,001 \\
        \bottomrule
    \end{tabular}
    \caption{Dataset statistics. Each cell represents the number of examples. K: thousand, M: million.}
    \label{tab:my_label}
\end{table}

\subsection{Experimental Setting}
We apply \textsc{Transformer}-\textit{Base}~\cite{vaswani2017attention} setting for all experiments.
The model dimensionality is 512, and 6 layers are engaged in both encoder and decoder side.
The inner-connection dimensionality for feedforward block is 2,048, and the number of heads in multi-head attention networks is 8.
We share the source embedding, target embedding and target softmax projection weight for En-Vi task, and share the latter two matrices for En-De.
We modify the learning rate schedule as:
$lr = 0.001 \cdot \min \left( \frac{t}{8000}, 1, (\frac{20000}{t}) ^ {0.5} \right),$
where $t$ denotes the current step.
Across all tasks, we determine the threshold $\tau$ as 1.0.

For both baseline and our model, En-Vi, En-De and Zh-En tasks take 50k, 150k and 200k updates, and each batch contains 4,096, 32,768 and 32,768 source tokens. The dropout ratio is determined as 0.3, 0.1 and 0.1, respectively.
All experiments are conducted over 4 NVIDIA V100 GPUs.
For each task, we choose the best model over dev set, defining beam size as 4, 4, 10 and decoding alpha as 1.5, 0.6 and 1.5, respectively.

\input{table_mt_overall_results}
\subsection{Experimental Results}
As shown in Table~\ref{tab:mt_overall_results}, vanilla model achieves best performance over all translation tasks.
However, replacing conventional attention networks with E-ATT does not lead to significant performance drop, with small decrease of 0.15$\sim$0.78 BLEU score.
Besides, after referring the statistics from Table~\ref{tab:cmos} and~\ref{tab:stats_model_operation}, our E-ATT module takes {34.10\%}/{33.83\%} energy of conventional ATT. %one, thus the energy consumption of entire attention network is 50.86\% of baseline.
These results reveal that, E-ATT can achieve competitive translation quality, and more importantly, significantly reduce the energy consumption of attention.
% In addition, we find that our model shows its advantage on energy cost than knowledge distillation methods, while maintaining competitive performance.~\footnote{See supplementary for comparison.}

\subsection{Ablation Study}
\input{table_ablation_pattern}
% To further study the influence of each component in our model, we 
We further conduct ablation experiments on En-Vi task.
As seen in Table~\ref{tab:ablation_pattern}, using discrete feature selection instead of linear transition slightly harms performance, with 0.61 BLEU score decrease.
Besides, replacing dot-product attention with $L_1$ distance does not significantly affect model performance. %, with only 0.07 BLEU score drop against baseline.
We can conclude that: 1) the performance gap between E-ATT and vanilla model mainly stems from the usage of discrete feature selection; and 2) $L_1$ distance can measure the similarity of vectorized representations and give modest performance compared to baseline. 
% Besides, the performance gap between E-ATT and vanilla model lies in the feature selection with discreteness.
% We further contribute the reason why discrete feature selection performs worse lies in the information loss of its discrete design~\cite{oda2017neural,qin2020binary}.

\section{Analyses}

\subsection{Hybrid Attention Networks}
\input{table_sa_overall_results}
We conduct a series of experiments involving hybrids of attention networks among vanilla ATT, Dense, RandInit, and E-ATT module in Table~\ref{tab:sa_overall_results}.
As shown, the conventional attention network performs the best among all models.
Our module performs well when served as either self-attention or cross-attention modules.
Besides, for all cases applying Dense/RandInit as cross-attention modules, models perform significantly worse, identical with the findings in~\newcite{tay2021synthesizer}.
% The performance drop is around 8.20$\sim$9.29 BLEU scores, which is identical with the findings in~\newcite{tay2021synthesizer}.
On the contrary, E-ATT module can give better performance with marginal performance drop comparing with baseline, indicating that E-ATT module is capable of providing adequate semantic alignments across languages for translation.
Besides, it is encouraging to see that our method works compatibly with other modules with marginal performance drop.

\subsection{Knowledge Distillation}
\input{figure_KD}
Knowledge distillation is a representative of model compression  approach~\cite{hinton2015distilling,tang2019distilling}. We further conduct experiments on ATT models with various dimensionalities compressed by knowledge distillation. 
 Figure~\ref{fig:mt_energy_cost_model_scale} shows the energy consumption and performance of different models with modified dimensionality $d$. 
As seen, by accumulatively halving $d$ from 512, both ATT and E-ATT progressively loses the quality. %, and finally diverge when $d$ is 32. 
%Besides, E-ATT saves around 99.45\% energy compared to baseline to achieve comparable performance against ATT.
%This proves the advantage of our 
However, E-ATT can give a better trade-off between model performance and energy consumption than knowledge distillation methods.

\subsection{Case Study}
\begin{figure}
    \centering
    \includegraphics[width=0.85\columnwidth]{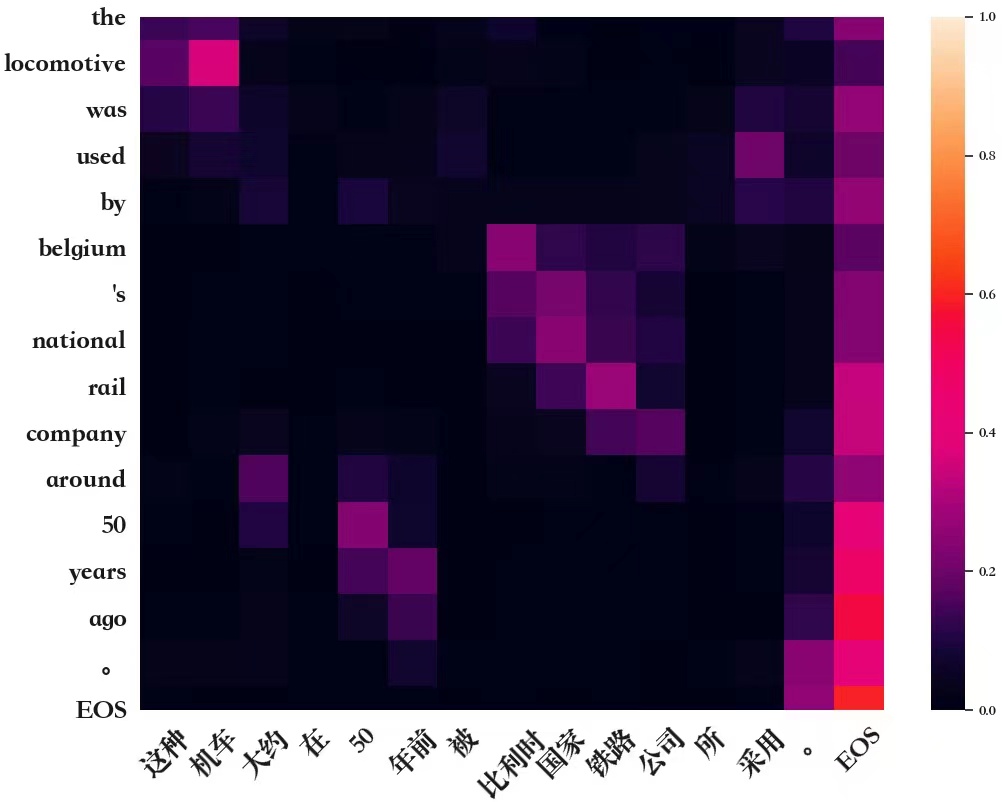}
    \caption{Case study from WMT'17 Zh-En dev set. E-ATT performs well on cross-lingual alignments.}
    \label{fig:my_label}
\end{figure}
We visualize the averaged attention values over one case from WMT'17 Zh-En dev set.
As seen, our model can give good aligned information, where preposition phrase "around 50 years ago" is arranged at the end of sentence in English, while its aligned phrase is at the front in Chinese.
This reveals that, our E-ATT can perform well on modeling the cross-lingual alignments.

\subsection{Binarization Statistics}
\input{figure_binary_ratio}
We further collect the ratio of nonzero values $\rho$ for each attention module in Figure~\ref{fig.binarization.ratio}, we can see that it increases with the number of encoder layers, denoting that more information is arranged into attentive calculation at higher layer of source side.
However, for decoder E-ATT, the ratio meets its peak at middle layers, revealing that decoder E-ATT are most active at the middle term of semantic processing.
Interestingly, ratio in the query of cross-attention modules, which align source and target semantics, is higher for the layer closer to output.
As the binarized key representation of each cross-attention module is equivalent, higher ratio of nonzero values in query representation means that, E-ATT at higher decoder layer provides more information for cross-lingual alignments, thus enrich the information for translation.
% Above all, we believe that our model helps investigate the interpretability, as the discrete values can reveal more intuitive statistics for semantic processing.

\section{Discussion and Conclusion}
In this paper, we empirically investigate the high energy-consumption problem in ATT. We argue that the alignment modeling procedure can be achieved with additions other than multiplications, thus reducing the energy costs. Extensive analyses suggest that: 1) Binarized representations marginally harm the feature extraction procedure; and 2) $L_1$ distance can be efficiently exploited to measure alignment among queries and keys. Compared to baseline, our approach can yield considerable quality of translations, and significantly save energy in attention mechanism. 
Although we have shown the superiority of E-ATT, considering the whole \textsc{Transformer} block\footnote{A \textsc{Transformer} block consists of a multi-head attention layer and a feed-forward layer. Please refer to Appendix~\ref{appendix.energy_ratio_calculation} for more calculation details.}, the use of E-ATT brings 17\% energy reduction. We hope this work can attract more researches on energy-efficient models. It is worth to further design techniques that reduce the energy cost of other modules in \textsc{Transformer}. 

\section*{Acknowledgements}
The authors would like to send great thanks to all reviewers and meta-reviewer for their insightful comments.
This work was supported in part by the Science and Technology Development Fund, Macau SAR (Grant No. 0101/2019/A2), the Multi-year Research Grant from the University of Macau (Grant No. MYRG2020-00054-FST), and Alibaba Group through Alibaba Research Intern Program.
% Special thanks to Jun Xie who offered valuable comments during paper revision.

%and 3) Binarization ratio of E-ATT can give more intuitive and detailed description for model interpretability.

% As our approach is transparent to model architectures and tasks, it is interesting to examine our method in other NLP tasks. Another promising direction is to optimize other modules in advanced Transformer from the energy-friendly aspects.
% Besides, it is interesting to examine our method in other NLP tasks, and optimize other modules besides ATT.

\bibliographystyle{acl_natbib}
\bibliography{anthology}

% \newpage

\appendix

\section{Energy Ratio Calculation}
\label{appendix.energy_ratio_calculation}
%As shown in Table~\ref{tab:stats_model_operation} and \ref{tab:mt_overall_results}, we collect the statistics on the number of additive/multiplicative operations. 

We calculate the energy cost following the common practice~\cite{chen2020addernet,you2020shiftaddnet}. 
Note that, we follow suggestions in \newcite{song2021addersr} to omit the energy calculation of activate functions, such as relu and softmax,  because they are specially designed over some modern AI chips, which requires far less energy than additive operation. 
%In this paper, we omit the energy required by activations for simplicity.

For the common case where input $\mathbf{X} \in \mathbb{R}^{l_1 \times d}$ is projected into $\mathbf{Q} \in \mathbb{R}^{l_1 \times d}$ with $\mathbf{W}\in \mathbb{R}^{d \times d}$:
\begin{align}
    \mathbf{Q} = \mathbf{X}\mathbf{W}^{\top} + b, %\mathbf{W}^{\top} \in \mathbb{R}^{d \times d}, b \in \mathbb{R}^{1 \times d},
\end{align}
the number of multiplicative operations is $l_1 \times d \times d + l_1 \times d = l_1d^2 + l_1d$.
For the number of additive operations, it is also $l_1d^2 + l_1d$.
Specially, if $d \gg 1$, we can omit the term $l_1d$ for simplicity.

We calculate the energy cost from three levels:
the alignment calculation which is used to measure the attention score, the whole attention model which is the core of our work, and the widely used \textsc{Transformer} \cite{vaswani2017attention} block, which contains a multi-head attention layer and a feedforward layer. We simply set the sequence length of inputs $\mathbf{X}$ and $\mathbf{Y}$ to $l$. Then: 
%Following this, $a$ equals to the length of input sentence $l$, $b$ and $c$ both equal to the dimension of model hidden states $d$.
\paragraph{Alignment Calculation} Two linear prjections are arranged to obtain query and key representations, yielding $2ld^2$ additive/multiplicative operations. Both query and key representations are used to derive attention logits. In dot-product, the number of required multiplicative operations is $l \times d \times l = l^2d$. The total numbers of additive/multiplicative operations are both $2ld^2 + l^2d$.  
\paragraph{Attention Model} In order to obtain value representations, attention model requires $ld^2$ additive/multiplicative operations.
    % Both representations are arranged for dot-product matrix multiplications to derive the logits before attention weights after splitted into $h$ heads (Equation~\ref{equ.Qsplithead}\&\ref{equ.KVsplithead}), where the number of required multiplicative operations is $h \times l \times k_{h} \times l = hl^2k_{h} = l^2d$.
    Besides, applying weighted sum over value representations with attention weights requires $l^2d$ multiplicative/additive operations.
    Overall, the numbers of multiplicative/additive operations in the whole attention model are $3ld^2 + 2l^2d$.
\paragraph{\textsc{Transformer} Block}  Although representations in multi-head attention are splitted into $h$ heads, in which dimension is $d_h$ ($d = h \times d_h$), the number of multiplicative/additive operations is also $h \times l \times d_h \times l = l^2hd_h = l^2d$. There are one output linear transition for the output of multiple heads and two additional linear transitions in feedforward layer, resulting in $ld^2 + 4ld^2 + 4ld^2 = 9ld^2$ additional additive/multiplicative operations. The overall multiplication operations in a \textsc{Transformer} Block is $9ld^2 + 3ld^2 +2l^2d = 12ld^2 + 2l^2d$.

\input{table_appendix_energy_cost}
Same as the steps above, we can calculate the required energy consumption of other modules. For example, considering our proposed E-ATT, the numbers of additive and multiplicative operations are $2ld + l^d$ and $0$ in alignment calculation, $ld^2+2ld+2l^2d$  and $ld^2 + l^2d$  in attention model,  $10ld^2 + 2ld + 2l^2d$ and $10ld^2 + l^2d$ in \textsc{Transformer} block. 

In this way, we can calculate the ratio of energy cost referring to the statistics in Table~\ref{tab:cmos}.
For example, the ratio between E-ATT and conventional attention model on ASIC chip is:

\small
\begin{align}
    \Delta_A = & \frac{0.9(ld^2 + 2ld  + 2l^2d) + 3.7(ld^2 + l^2d)}{0.9(3ld^2 + 2l^2d) + 3.7(3ld^2 + 2l^2d)} %\\
    %= & \frac{1.8 + 5.5l + 4.6d}{13.8d + 9.2l}.
\end{align}
\normalsize
Similarly, that on FPGA chip is:
\small
\begin{align}
    \Delta_F = & \frac{0.4(ld^2 + 2ld + 2l^2d)+18.8(ld^2 + l^2d)}{0.4(3ld^2 + 2l^2d) + 18.8(3ld^2 + 2l^2d)} %\\
    %= & \frac{0.8 + 19.6l + 19.2d}{57.6d + 38.4l}.
\end{align}
\normalsize
For the IWSLT'15 En-Vi task,  with $d$ being 512, the averaged length of dataset is $\bar{l}=22$. We can get the result $\Delta_A$ = 34.09\%, $\Delta_F$ = 33.83\%, thus the energy reduction ratio is $1 - \Delta_A$ = 65.91\%, $1 - \Delta_F$ = 66.17\%.
Similarly, energy consumption ratios at the level of alignment calculation are 99.55\% and 99.95\%. Those of \textsc{Transformer} block are 83.17\% and 83.10\%, respectively.

%\newpage

\end{document}

% \maketitle

\appendix

\section{Dataset Preprocessing}
In this paper we evaluate our approach with three widely used machine translation datasets: IWSLT'15 English -Vietnamese (En-Vi), WMT'14 English - German (En-De) and WMT'17 Chinese - English (Zh-En).
All datasets are segmented into subwords by byte-pair encoding ~\citep[BPE,][]{sennrich2016neural} with 32k merge operations.
Specially, for the former two tasks, we apply joint BPE for both source and target languages.
All datasets are modified into truecase format with~\href{https://github.com/mosesdecoder/}{mosesdecoder} by training truecase models upon train set.

\begin{table}[h]
    \centering
    \begin{tabular}{c||ccc}
        \hline
        Dataset & Train & Dev & Test \\
        \hline
        \hline
        En-Vi & 13.3K & 1,553 & 1,268 \\
        En-De & 4.50M & 3,000 & 3,003 \\
        Zh-En & 20.6M & 2,002 & 2,001 \\
        \hline
    \end{tabular}
    \caption{Dataset statistics. Each cell represents the number of examples. K: thousand, M: million.}
    \label{tab:my_label}
\end{table}

\section{Experimental Setting}
We apply \textsc{Transformer}-\textit{Base}~\cite{vaswani2017attention} setting for all experiments.
The model dimensionality is 512, and 6 layers are engaged in both encoder and decoder side.
The inner-connection dimensionality for feedforward block is 2,048, and the number of heads in multi-head attention networks is 8.
We share the source embedding, target embedding and target softmax projection weight for En-Vi task, and share the latter two matrices for En-De.
We modify the learning rate schedule as:
$lr = 0.001 \cdot \min \left( \frac{t}{8000}, 1, (\frac{20000}{t}) ^ {0.5} \right),$
where $t$ denotes the current step.
Across all tasks, we determine the threshold $\tau$ as 1.0.

For both baseline and our model, En-Vi, En-De and Zh-En tasks take 50k, 150k and 200k updates, and each batch contains 4,096, 32,768 and 32,768 source tokens. The dropout ratio is determined as 0.3, 0.1 and 0.1, respectively.
All experiments are conducted over 4 NVIDIA V100 GPUs.
For each task, we choose the best model over dev set, defining beam size as 4, 4, 10 and decoding alpha as 1.5, 0.6 and 1.5, respectively.

\section{Binarization Statistics}
We further collect the ratio of nonzero values $\rho$ for each attention module in Figure~\ref{fig.binarization.ratio}, we can see that it increases with the number of encoder layers, denoting that more information is arranged into attentive calculation at higher layer of source side.
However, for decoder E-ATT, the ratio meets its peak at middle layers, revealing that decoder E-ATT tends to focus on target semantics at the middle term of semantic processing.
Interestingly, ratio in the query of cross-attention modules, which align source and target semantics, is higher for the layer closer to output.
As the binarized key representation of each cross-attention module is equivalent, higher ratio of nonzero values in query representation means that, E-ATT at higher decoder layer provides more information for cross-lingual alignments, thus enrich the information for translation.
% Above all, we believe that our model helps investigate the interpretability, as the discrete values can reveal more intuitive statistics for semantic processing.

\input{figure_binary_ratio}

\section{Case Study}
We visualize the averaged attention values over one case from WMT'17 Zh-En dev set.
As seen, our model can give good aligned information, where preposition phrase "around 50 years ago" is arranged at the end of sentence in English, while its aligned phrase is at the front in Chinese.

\begin{figure}
    \centering
    \includegraphics[width=\columnwidth]{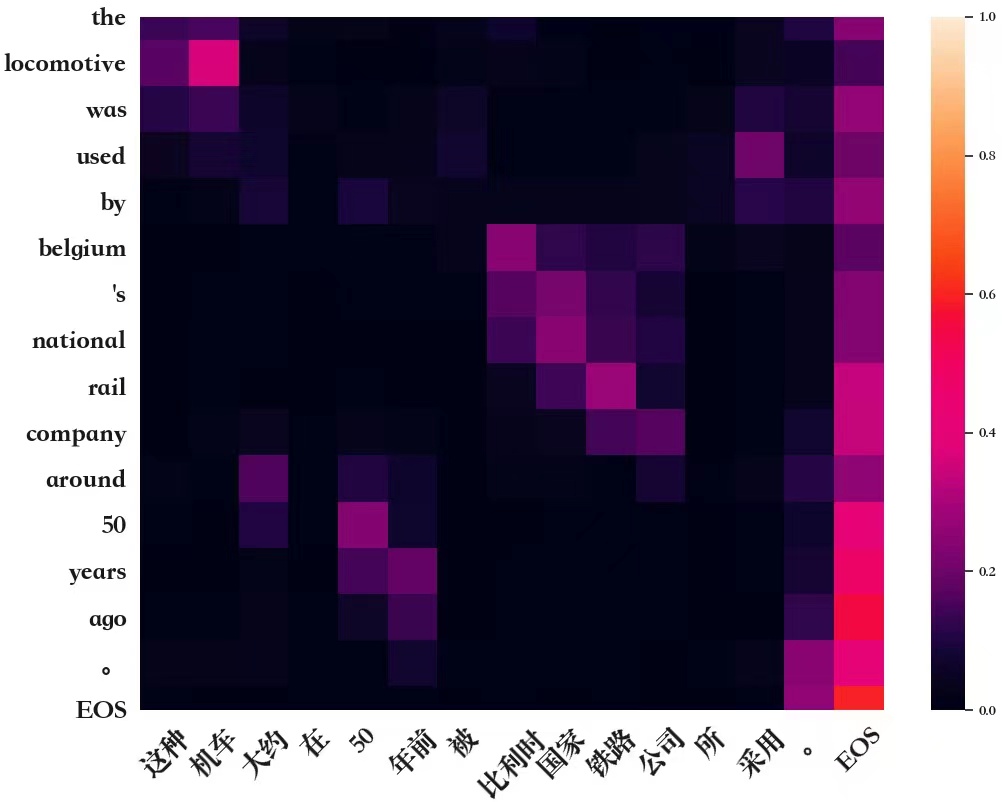}
    \caption{Case study from WMT'17 Zh-En dev set.}
    \label{fig:my_label}
\end{figure}

% \section{Knowledge Distillation}
% In order to compare the model performance with proposed E-ATT and distilled baseline model~\cite{hinton2015distilling,tang2019distilling}, we further conduct a group of experiments on the various dimensionalities inside ATT.
% As in Figure~\ref{fig:mt_energy_cost_model_scale}, we simulate the energy consumption of each model with modified dimensionality $k$, and conduct the relationship between corresponding energy cost and performance.
% As seen, by accumulatively halving the dimensionality of model $d$ from 512, \textsc{Transformer} significantly losses the performance, and finally diverge when $d$ is 32.
% In comparison, the proposed E-ATT can save almost 15 times energy than that of baselines with knowledge distillation, verifying the superiority of our approach. % on both to obtain comparative performance.
%This demonstrates that our approach can significantly reduce energy consumption, and still be trainable given quite small energy costs against baseline model.

% \input{figure_KD}

\bibliographystyle{acl_natbib}
\bibliography{anthology}

%\appendix

%% file: table_cmos_energy.tex
% \begin{table}[t]
%     \centering
%     \begin{tabular}{l||c|c}
%     \hline
%         Operation (FP32) & ASIC & FPGA \\
%     \hline
%     \hline
%       Addition & 0.9 & 0.4 \\
%       Multiplication & 3.7 & 18.8 \\
%     \hline
%       Ratio & \textbf{4.1} & \textbf{47.0} \\
%     \hline 
%     \end{tabular}
%     \caption{Energy cost ($pJ$) of addition/multiplication operation on ASIC/FPGA hardware from~\newcite{you2020shiftaddnet}, and the ratio of multiplication energy cost to addition. $1 pJ = 10^{-12} Joule$. A multiplicative operation costs far more energy than additive one.}
%     \label{tab:cmos}
% \end{table}

% % Energy cost ($pJ$) of addition/multiplication operation on ASIC/FPGA hardware from~\newcite{you2020shiftaddnet}, and the ratio of multiplication energy cost to addition. $1 pJ = 10^{-12} Joule$. A multiplicative operation costs far more energy than additive one.

\begin{table}[t]
    \centering
    \small
    % \begin{tabular}{l||rr}
    \begin{tabular}{lrr}
    % \hline
    \toprule
        \textbf{Operation} (FP32) & \textbf{ASIC} & \textbf{FPGA} \\
    % \hline
    % \hline
    \midrule
      Addition & 0.9~~ & 0.4~~ \\
      Multiplication & 3.7~~ & 18.8~~ \\
    % \hline
    %   Ratio & \textbf{4.1} & \textbf{47.0} \\
    % \hline 
    \bottomrule
    \end{tabular}
    \caption{Energy cost ($10^{-12} Joule$) of addition and multiplication on ASIC/FPGA chips \cite{you2020shiftaddnet}. The representative of ASIC chip is Google's TPU, while Microsoft cloud employs FPGA chips.} % As seen, multiplication requires far more energy than addition.}% }
    \label{tab:cmos}
\end{table}

%% file: table_calls_of_operators.tex
\begin{table}[t]
    \centering
    \small
    \scalebox{0.9}
    {
        \begin{tabular}{lcrcr}
        \toprule
            \multirow{2}{*}{\textbf{Model}} & \multicolumn{2}{c}{\textbf{Alignment}}   &\multicolumn{2}{c}{\textbf{Attention}}  \\
            \cmidrule(lr){2-3} \cmidrule(lr){4-5}
             & \# \textbf{mul} & $\Delta_A(\%)$ & \# \textbf{mul} & $\Delta_A(\%)$ \\
        % \hline
        \midrule
            Vanilla & $2ld^2+l^2d$ &100.00 & $3ld^2+2l^2d$ & 100.00~ \\
            Dense & $ld^2+l^2d$  & 51.10  & $2ld^2+2l^2d$ & 67.59~ \\
            RandInit & $0$ & 0.00 & $ld^2+l^2d$ & 33.80~ \\
            %Factorized Dense & $H_A(F_A(X))* H_B(F_B(X))$ %& $d^2l+d(k_A+k_B)l$ & $d^2l+d(k_A+k_B)l$  \\
          % Factorized Random & $R_1R_2^T$ & $kl^2$ &  %$kl^2$\\
            % Only $l_1$ norm & $-|| (X-F_Q)-(X-F_K) ||_1$ & $2d^2l+dl^2$ & 0\\
        \cdashline{1-5}\noalign{\vskip 0.1ex}
            E-ATT & $0$ & 0.44 & $ld^2 + l^2d$ & 34.10~ \\
        
        \bottomrule
        \end{tabular}
    }
    % \caption{Calls of addition (add) / multiplication (mul), and energy consumption ratio over ASIC chip \textcolor{red}{($\Delta$)} to derive attention scores in vanilla ATT~\cite{vaswani2017attention}, Dense~\cite{tay2021synthesizer}, RandInit~\cite{tay2021synthesizer}, and our model \textcolor{red}{following} \textsc{Transformer}-\textit{Base} setting.
    \caption{Calls of multiplication (mul) and energy consumption ratio on ASIC chip ($\Delta_A$) of vanilla ATT, Dense, RandInit, and our model. ``Alignment'' and ``Attention'' indicate the statistics are conducted at the level of alignment calculation and the whole attention model, respectively.   % to derive attention scores in by referring to Table~\ref{tab:cmos}.
    % Results are conducted over \textsc{Transformer}-\textit{Base} setting.
    % Results are conducted when each attention module receive input $\mathbf{X} \in \mathcal{R}^{l_1 \times d}, \mathbf{Y} \in \mathcal{R}^{l_2 \times d}$. Each model gives query and key representations $\mathbf{Q} \in \mathcal{R}^{l_1 \times k}, \mathbf{K} \in \mathcal{R}^{l_2 \times k}$, and calculates alignment scores $\mathbf{A} \in \mathcal{R}^{l_1 \times l_2} $ for each head. 
    $l$ and $d$ are sequential length and model size. Please refer to Appendix~\ref{appendix.energy_ratio_calculation} for more details.}
    \label{tab:stats_model_operation}
\end{table}

%% file: table_mt_overall_results.tex
\begin{table*}[t]
    \centering
    \scalebox{0.95}{
        \begin{tabular}{lcccrr}
        \toprule
            \textbf{Attention mechanism} & \textbf{En-Vi} & \textbf{En-De} & \textbf{Zh-En} & {\textbf{ASIC}} (\%) & {\textbf{FPGA} (\%)} \\
        \midrule
            Vanilla & 30.26 $\pm$ 0.07 & 27.60 $\pm$ 0.04 & 24.28 $\pm$ 0.08 & 100.00~~~ & 100.00~~~ \\
            E-ATT & 29.48 $\pm$ 0.08 & 27.45 $\pm$ 0.04 & 24.23 $\pm$ 0.06 & \textbf{34.10}~~~ & {\textbf{33.83}}~~~ \\
        \bottomrule
        \end{tabular}
    }
    \caption{Averaged BLEU scores (\%) upon test set on IWSLT'15 En-Vi, WMT'14 En-De and WMT'17 Zh-En tasks over 5 independent runs. E-ATT gives comparable results against conventional ATT, reducing the energy cost at 65.90\%/66.17\% on ASIC/FPGA chip in attention procedure.
    % Since the energy cost for a specific module is difficult to be empirically evaluated, we report the theoretical values following the common practice~\cite{chen2020addernet,you2020shiftaddnet}.
    Since the energy cost is difficult to be empirically evaluated, we report the theoretical values following the common practice~\cite{chen2020addernet,you2020shiftaddnet}.}
    \label{tab:mt_overall_results}
\end{table*}

%% file: table_ablation_pattern.tex
\begin{table}[t]
    \centering
    \begin{tabular}{lc}
    \toprule
        \textbf{Model} & \textbf{BLEU} (\%) \\
    \midrule
        Vanilla & \textbf{28.12} \\
    % \cdashline{1-2}
        ~~~Replace with discrete selection & 27.51 \\
        ~~~Replace with $L_1$ distance & 28.05 \\
    % \hline
    \cdashline{1-2}
        E-ATT & 27.45 \\
        % $L_1$ distance & 28.10 \\
        % E-ATT & 27.45 \\
        % ~~~- binaried & 28.05 \\
        % ~~~- negative  & 27.04 \\
        % ~~~- binaried + negative & 27.66\\
        % ~~~- $L_1$ norm + negative & 27.45 \\
    \bottomrule 
    \end{tabular}
    \caption{Model performance with component replacements over En-Vi dev set. Using $L_1$ distance as similarity measurement does not harm model performance.}
    \label{tab:ablation_pattern}
\end{table}

%% file: table_sa_overall_results.tex
% \begin{table}[t]
%     \centering
%     \begin{tabular}{c|c||c}
%     \hline
%         Self-Attn & Crs-Attn & BLEU (\%) \\
%     \hline
%     \hline
%         vanilla & vanilla & \textbf{28.12} \\
%         Dense & Dense & 19.43 \\
%         RandInit & RandInit & 18.83 \\
%         % vocab size: 46152
%     \cdashline{1-3}
%         A$^3$N & A$^3$N & 27.45 \\
%     \hline
%         Dense & vanilla & 27.48 \\
%         RandInit & vanilla & 27.36 \\
%         vanilla & Dense & 19.92 \\
%         vanilla & RandInit & 19.31 \\
%     \cdashline{1-3}
%     % \hline 
%         vanilla & A$^3$N & 27.72 \\
%         Dense & A$^3$N & 27.60 \\
%         RandInit & A$^3$N & 27.48 \\
        
%         A$^3$N & vanilla & 28.08 \\
%         A$^3$N & Dense & 19.85 \\
%         A$^3$N & RandInit & 19.67 \\
%     \hline
%     \end{tabular}
%     \caption{BLEU score (\%) of different model hybrids with modifying self-attention (Self-Attn) and cross-attention (Crs-Attn) network upon IWSLT'15 En-Vi dev set. The \textsc{Transformer} model gives the best performance, and our proposed model can achieve 27.45 BLEU score, dropping 0.67 BLEU score against best baseline model. Besides, our A$^3$N can achieve good performance when applied as cross attention modules, whereas Dense or RandInit can not.}
%     \label{tab:sa_overall_results}
% \end{table}

\begin{table}[t]
    \centering
    % \small
    \scalebox{0.95}{
        \begin{tabular}{ccccc}
        \toprule
            % \diagbox[width=5em,trim=l]{Self}{Crs} & vanilla & Dense & RI & A$^3$N \\
            - & Vanilla & Dense & RandInit & E-ATT \\
        % \hline
        \midrule
            Vanilla & \textbf{28.12} & 19.92 & 19.31 & 27.72 \\
            Dense & 27.48 & 19.43 & 19.21 & 27.60 \\
            RandInit & 27.36 & 18.98 & 18.83 & 27.48 \\
            % vocab size: 46152
        % \cdashline{1-5}
            E-ATT & 28.08 & 19.85 & 19.67 & 27.45 \\
        % \hline
        \bottomrule
        \end{tabular}
    }
    \caption{BLEU score (\%) of different model hybrids with modifying self-attention (horizontal) and cross-attention (vertical) network upon En-Vi dev set. 
    % E-ATT achieves 27.45 BLEU score, dropping 0.67 BLEU score against best baseline model. 
    % Besides, E-ATT can achieve good performance when applied as cross attention modules, whereas Dense or RandInit can not.
    E-ATT can achieve good performance when applied as cross-attention modules, whereas Dense or RandInit can not.}
    \label{tab:sa_overall_results}
\end{table}

%% file: figure_KD.tex
\begin{figure}[t]
    \centering
    \begin{tikzpicture}
    \pgfplotsset{set layers}
    
    \begin{axis}[
        height=0.6 * \columnwidth,
        width=\columnwidth,
        title={},
        xlabel={Energy consumption ratio (\%)},
        ytick pos=left,
        xtick pos=bottom,
        scaled x ticks=true,
        % xtick scale label code/.code={},
        % xmode=log,
        ylabel={BLEU (\%)},
        xmin=-5, xmax=105,
        ymin=19, ymax=29,
        xtick={0, 20, 40, 60, 80, 100},
        ytick={19, 24, 29},
        % ymajorgrids=true,
        % xmajorgrids=true,
        grid=major,
        grid style=dashed,
        legend cell align=left,
        legend style={
            at={(0.95, 0.05)},
            anchor=south east,
            font=\scriptsize,
			legend columns=1}
    ]
    
    \addplot[
        color=blue,
        thick,
        mark=triangle*,
        ]
        coordinates{
            (100, 28.12)
            (50, 27.22)
            (25, 25.52)
            (12.5, 21.08)
        };
    \addlegendentry{ATT}
    
    \addplot[
        color=black,
        thick,
        mark=x,
        ]
        coordinates{
            (34.09, 27.45)
            (17.05, 27.32)
            (8.53, 26.42)
            (4.28, 20.07)
        };
    \addlegendentry{E-ATT}
    
    \addplot[
        color=gray,
        ] 
        coordinates{
        (-10, 27.45)
        (110, 27.45)
    };
    
    \end{axis}
    \end{tikzpicture}
        
    \caption{Performance and energy consumption of different models with knowledge distillation on En-Vi dev set. We regard the energy consumption of ATT baseline as 1, and accumulatively halve the dimensionality of model till untrainable (from 512 to 64).
    Energy consumption is estimated over ASIC.
    % E-ATT performs better following the same ratio of energy costs.
    E-ATT requires far less energy to meet up the baseline performance.}
    \label{fig:mt_energy_cost_model_scale}
\end{figure}

%% file: figure_binary_ratio.tex
\begin{figure}
    \centering
    \begin{tikzpicture}
        \pgfplotsset{set layers}
        \pgfplotsset{every x tick label/.append style={font=\small}}
        \pgfplotsset{every y tick label/.append style={font=\small}}
        \begin{axis}[
            ybar,
            bar width=0.02 * \columnwidth,
            height=0.5 * \columnwidth,
            width=0.95 * \columnwidth,
            title={},
            xlabel={Layer},
            ytick pos=left,
            xtick pos=bottom,
            % scaled x ticks=true,
            % xtick scale label code/.code={},
            ylabel={Ratio(\%)},
            % domain=0:4,
            % xmin=0, xmax=2,
            ymin=0, ymax=60,
            symbolic x coords={1, 2, 3, 4, 5, 6},
            xtick=data,
            enlarge x limits=0.1,
            ytick={0, 20, 40, 60},
            yticklabels={0, 20, 40, 60},
            ymajorgrids=true,
            grid style=dashed,
            legend cell align=left,
            legend style={
                at={(0.50, 1.08)},
                anchor=south,
                font=\scriptsize,
    			legend columns=4},
		    every axis plot/.append style={thick},
    		]
    	
    	\addplot[
        	    color=black!40!green,
        	    fill=black!40!green,
        	    fill opacity=0.3,
        	]
        	coordinates {
        	    (1, 5.90)
        	    (2, 5.85)
        	    (3, 15.03)
        	    (4, 17.40)
        	    (5, 17.69)
        	    (6, 17.13)
        	};
        	\addlegendentry{Enc-Self}
    	
    	\addplot[
        	    color=blue,
        	    fill=blue,
        	    fill opacity=0.3,
        	]
        	coordinates {
        	    (1, 20.17)
        	    (2, 23.73)
        	    (3, 28.08)
        	    (4, 28.20)
        	    (5, 28.30)
        	    (6, 23.88)
        	};
        	\addlegendentry{Dec-Self}
        	
        \addplot[
                color=red,
        	    fill=red,
        	    fill opacity=0.3,
            ]
            coordinates {
        	    (1, 33.87)
        	    (2, 36.83)
        	    (3, 45.09)
        	    (4, 47.26)
        	    (5, 54.99)
        	    (6, 54.73)
        	};
        	\addlegendentry{Dec-Crs-Query}
        	
        \addplot[
            color=orange,
            fill=orange,
            fill opacity=0.3,
            ]
            coordinates {
                (1, 24.20)
        	    (2, 24.20)
        	    (3, 24.20)
        	    (4, 24.20)
        	    (5, 24.20)
        	    (6, 24.20)
            };
            \addlegendentry{Dec-Crs-Key}
    		
	    \end{axis}
	    
    \end{tikzpicture}
    \caption{Ratio of nonzero values in the representations of E-ATT. Enc-Self: encoder self-attention ; Dec-Self: decoder self-attention;
    Dec-Crs-Query/Key: query/key representation for decoder cross-attention. 
    % Ratio of nonzero values in decoder self-attention modules meet its peak at the medium of processing, while that goes larger in deeper encoder E-ATT module. 
    Query representations in cross-attention are the most active.}
    % \textcolor{red}{What does higher ratio mean? Activated? Model requires more information for alignments?}
    
    \label{fig.binarization.ratio}
\end{figure}
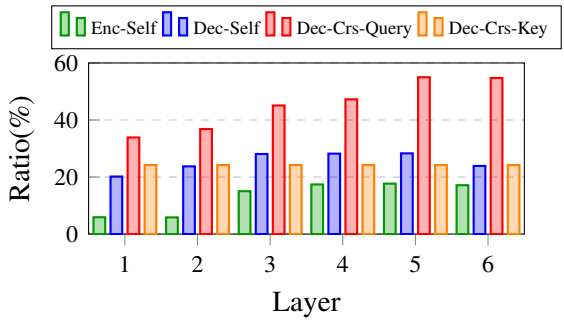

%% file: table_appendix_energy_cost.tex
\begin{table}[t]
    \centering
    \begin{tabular}{l:rr}
    \toprule
        Calculation Level & $\Delta_A$ & $\Delta_F$ \\
    \midrule
        Alignment Calculation & 0.45 & 0.05 \\
        Attention Model & 34.09 & 33.83 \\
        \textsc{Transformer} Block & 83.17 & 83.10 \\
    \bottomrule
    \end{tabular}
    \caption{Energy consumption ratio (\%) at each calculation level on En-Vi translation task compared to conventional \textsc{Transformer} design.}
    \label{tab:energy_cost}
\end{table}